%% file: deid.tex
\documentclass[11pt,letterpaper]{article}
\usepackage{classes}
\usepackage{times}
\usepackage{latexsym}

\usepackage{multicol}
\usepackage{abstract}
\usepackage{amsmath}
\usepackage{graphicx}
\usepackage{latexsym}
\usepackage{diagbox}
\usepackage{amssymb}

\usepackage{booktabs}
\usepackage{float}

\usepackage{adjustbox}

\usepackage[table]{xcolor}
\usepackage{array}
\usepackage{multirow}
\usepackage{tabularx}
\usepackage{multicol}
\usepackage{multirow}
\usepackage{diagbox}
\usepackage{amsmath}
\usepackage{booktabs}
\usepackage{float}
\restylefloat{table}
\usepackage[font=footnotesize,labelfont=bf]{caption}
\usepackage{enumitem}
\usepackage{makecell}
\newcolumntype{x}[1]{>{\centering\arraybackslash}p{#1}}

\usepackage{hyperref}
\hypersetup{
  hidelinks % Remove visible links altogether
}

\DeclareMathOperator{\argmax}{argmax}

\usepackage{pbox}

\naaclfinalcopy
\def\naaclpaperid{***}

\title{De-identification of Patient Notes with Recurrent Neural Networks}

  \author{\hspace{2cm}Franck Dernoncourt\thanks{\hspace{1mm}These authors contributed equally to this work.}~, Ji Young Lee\footnotemark[1]~, Peter Szolovits \\ 
	    \hspace{2cm}MIT\\
	    \hspace{2cm}Cambridge, MA, USA\\
	    \hspace{2cm}{\tt \{francky,jjylee,psz\}@mit.edu}
	  \And
	 \hspace{2cm}{\"O}zlem Uzuner\\
  	\hspace{2cm}SUNY Albany\\
  	\hspace{2cm}Albany, NY, USA\\
  \hspace{2cm}{\tt ouzuner@albany.edu}  
  }

\date{}

\begin{document}

   \vspace{-1cm}
\twocolumn[{
 \begin{@twocolumnfalse}
 \centering
   \maketitle
    \input{abstract.tex}
 \end{@twocolumnfalse}
 }]
\saythanks

\input{introduction.tex}

\input{method.tex}

\input{results.tex}

\input{conclusion.tex}

\newpage
\section*{Funding}
The project was supported by Philips Research. The content is solely the responsibility of the authors and does not necessarily represent the official views of Philips Research.

\section*{Acknowledgments}

We warmly thank Michele Filannino, Alistair Johnson, and Tom Pollard for their helpful suggestions and technical assistance.

\bibliography{deid}
\bibliographystyle{classes}

\end{document}

%% file: abstract.tex
\begin{abstract}

\noindent\textbf{Objective}: 
Patient notes in electronic health records (EHRs) may contain critical information for medical investigations. However, the vast majority of medical investigators can only access de-identified notes, in order to protect the confidentiality of patients. In the United States, the Health Insurance Portability and Accountability Act (HIPAA) defines 18 types of protected health information (PHI) that needs to be removed to de-identify patient notes. Manual de-identification is impractical given the size of EHR databases, the limited number of researchers with access to the non-de-identified notes, and the frequent mistakes of human annotators. A reliable automated de-identification system would consequently be of high value.\vspace{0.2cm}

\noindent\textbf{Materials and Methods}:
We introduce the first de-identification system based on artificial neural networks (ANNs), which requires no handcrafted features or rules, unlike existing systems. We compare the performance of the system with state-of-the-art systems on two datasets: the i2b2 2014 de-identification challenge dataset, which is the largest publicly available de-identification dataset, and the MIMIC de-identification dataset, which we assembled and is twice as large as the i2b2 2014 dataset.\vspace{0.2cm}

\noindent\textbf{Results}:
Our ANN model outperforms the state-of-the-art systems. 
It yields an F1-score of 97.85 on the i2b2 2014 dataset, with a recall 97.38 and a precision of 97.32, and an F1-score of 99.23 on the MIMIC de-identification dataset, with a recall 99.25 and a precision of 99.06. \vspace{0.2cm}

\noindent\textbf{Conclusion}:
Our findings support the use of ANNs for de-identification of patient notes, as they show better performance than previously published systems while requiring no feature engineering.

\vspace{0.5cm}

\end{abstract}

%% file: introduction.tex
\section{Introduction and related work}

In many countries such as the United States, medical professionals are strongly encouraged to adopt electronic health records (EHRs) and may face financial penalties if they fail to do so~\cite{desroches2013some,wright2013early}. The Centers for Medicare \& Medicaid Services have paid out more than \$30 billion in EHR incentive payments to hospitals and providers who have attested to meaningful use as of March 2015. Medical investigations may greatly benefit from the resulting increasingly large EHR datasets. One of the key components of EHRs is patient notes: the information they contain can be critical for a medical investigation because much information present in texts cannot be found in the other elements of the EHR. However, before patient notes can be shared with medical investigators, some types of information, referred to as protected health information (PHI), must be removed in order to preserve patient confidentiality. In the United States, the Health Insurance Portability and Accountability Act (HIPAA)~\cite{office2002standards} defines 18 different types of PHI, ranging from patient names to phone numbers. Table~\ref{tab:phi_types} presents the exhaustive list of PHI types as defined by HIPAA.

The task of removing PHI from a patient note is referred to as de-identification, since the patient cannot be identified once PHI is removed.  De-identification can be either manual or automated. Manual de-identification means that the PHI are labeled by human annotators. There are three main shortcomings of this approach. First, only a restricted set of individuals is allowed to access the identified patient notes, thus the task cannot be crowdsourced. Second, humans are prone to mistakes. ~\cite{neamatullah2008automated} asked 14 clinicians to detect PHI in approximately 130 patient notes: the results of the manual de-identification varied from clinician to clinician, with recall ranging from 0.63 to 0.94.
\cite{douglass2005identification,douglas2004computer} reported that annotators were paid US\$50 per hour and read 20,000 words per hour at best. As a matter of comparison, the MIMIC dataset~\cite{goldberger2000physiobank,saeed2011multiparameter}, which contains data from 50,000 intensive care unit (ICU) stays, consists of 100 million words. This would require 5,000 hours of annotation, which would cost US\$250,000 at the same pay rate. Given the annotators' spotty performance, each patient note would have to be annotated by at least two different annotators, so it would cost at least US\$500,000 to de-identify the notes in the MIMIC dataset.

Automated de-identification systems can be classified into two categories: rule-based systems and machine-learning-based systems. Rule-based systems typically rely on patterns, expressed as regular expressions and gazetteers, defined and tuned by humans. They do not require any labeled data (aside from labels required for evaluating the system), and are easy to implement, interpret, maintain, and improve, which explains their large presence in the industry~\cite{chiticariu2013rule}. However, they need to be meticulously fine-tuned for each new dataset, are not robust to language changes (e.g., variations in word forms, typographical errors, or infrequently used abbreviations), and cannot easily take into account the context (e.g., ``Mr.\ Parkinson'' is PHI, while ``Parkinson's disease'' is not PHI). Rule-based systems are described in~\cite{berman2003concept,beckwith2006development,fielstein2004algorithmic,friedlin2008software,gupta2004evaluation,morrison2009repurposing,neamatullah2008automated,ruch2000medical,sweeney1996replacing,thomas2002successful}.

To alleviate some downsides of the rule-based systems, there have been many attempts to use supervised machine learning algorithms to de-identify patient notes by training a classifier to label each word as PHI or not PHI, sometimes distinguishing between different PHI types. Common statistical methods include decision trees~\cite{szarvas2006multilingual}, log-linear models, support vector machines~\cite{guo2006identifying,uzuner2008identifier,hara2006applying}, and conditional random fields~\cite{aberdeen2010mitre}, the latter being employed in most of the state-of-the-art systems. For a thorough review of existing systems, see ~\cite{meystre2010automatic,stubbs2015automated}.  All these methods share two downsides: they require a decent sized labeled dataset and much feature engineering. As with rules, quality features are challenging and time-consuming to develop.

Recent approaches to natural language processing based on artificial neural networks (ANNs) do not require handcrafted rules or features, as they can automatically learn
effective features by performing composition over tokens which are represented as vectors, often called token embeddings.
The token embeddings are jointly learned with the other parameters of the ANN. They can be initialized randomly, but can be pre-trained using large unlabeled datasets typically based on token co-occurrences~\cite{mikolov2013distributed,collobert2011natural,pennington2014glove}. The latter often performs better, since the pre-trained token embeddings explicitly encode many linguistic regularities and patterns.  As a result, methods based on ANNs have shown promising results for various tasks in natural language processing, such as language modeling~\cite{mikolov2010recurrent}, text classification~\cite{socher2013recursive,kim2014convolutional,blunsom2014convolutional,lee2016sequential}, question answering~\cite{weston2015towards,wang-nyberg:2015:ACL-IJCNLP}, machine translation~\cite{bahdanau2014neural,tamura2014recurrent,sundermeyer2014translation}, as well as named entity recognition~\cite{collobert2011natural,lample2016neural,labeau-loser-allauzen:2015:EMNLP}. A few methods also use vector representations of characters as inputs in order to either replace or augment token embeddings~\cite{kim2015character,lample2016neural,labeau-loser-allauzen:2015:EMNLP}.

\input{tables/phi_types}

Inspired by the performance of ANNs for various other NLP tasks, this article introduces the first de-identification system based on ANNs. Unlike other machine learning based systems, ANNs do not require manually-curated features, such as those based on regular expressions and gazetteers. We show that ANNs achieve state-of-the-art results on de-identification of two different datasets for patient notes, the i2b2 2014 challenge dataset and the MIMIC dataset.

%% file: tables/phi_types.tex
\begin{table*} [ht]
\footnotesize
\centering
\setlength\tabcolsep{6pt}
\setlength{\extrarowheight}{3pt}
\setlength{\arraycolsep}{5pt}
\begin{tabular}{|l|l|c|c|c|}
\hline
\textbf{PHI categories} & \textbf{PHI types} & \textbf{HIPAA} & \textbf{  i2b2  } & \textbf{MIMIC}  \\
\hline
AGE        & Ages $\ge$ 90                      & x & x & x\\
           & Ages $<$ 90                                          &   & x & \\
\hline
CONTACT    & Telephone and fax numbers                                      & x & x & x\\
           & Electronic mail addresses                                      & x & x & x\\
           & URLs or IP addresses*                                                           & x & x & x\\
\hline
DATE       & Dates (month and day parts)                                    & x & x & x\\
           & Year                                                           &   & x & x\\
           & Holidays                                                       &   & x & x\\
           & Day of the week                                                &   & x & \\
\hline
ID         & Social security numbers                                        & x & x & x\\
           & Medical record numbers                                         & x & x & x\\

           & Account numbers                                                & x & x & x\\
           & Certificate or license numbers                                 & x & x & x\\
           & Vehicle or device identifiers                                            & x & x & x\\

           & Biometric identifiers or full face photographic images*                                        & x & x & x\\

\hline
LOCATION   & Addresses and their components smaller than a state            & x & x & x\\
           & State                                                          &   & x & x\\
           & Country                                                        &   & x & x\\
           & Employers                                                      & x & x & x\\
           & Hospital name                                                  &   & x & x\\
           & Ward name                                                      &   &   & x\\
\hline
NAME       & Names of patients and family members                           & x & x & x\\
           & Provider name                                                  &   & x & x\\
\hline
PROFESSION & Profession                                                     &   & x & \\
\hline
\end{tabular}
\caption{PHI types as defined by HIPAA, i2b2, and MIMIC. PHI categories are defined in the i2b2 dataset. The PHI types marked with * do not appear in either dataset.}\label{tab:phi_types}
\end{table*}

%% file: method.tex
\section{Methods and materials}

We first present a de-identifier we developed based on a conditional random field (CRF) model in Section~\ref{sec:crf-system}. This de-identifier yields state-of-the-art results on the i2b2 2014 dataset, which is the reference dataset for comparing de-identification systems.
This system will be used as a challenging baseline for the ANN model that we will present in Section~\ref{sec:ann-system}.
The ANN model outperforms the CRF model, as outlined in Section~\ref{sec:results}.

\subsection{CRF model} \label{sec:crf-system}
In the CRF model, each patient note is tokenized and features are extracted for each token. During the training phase, the CRF's parameters are optimized to maximize the likelihood of the gold standard labels. During the test phase, the CRF predicts the labels. 
The performance of a CRF model depends mostly on the quality of its features. We used a combination of n-gram, morphological, orthographic, and gazetteer features. 
These are similar to features used in the best-performing CRF-based competitors in the i2b2 challenge~\cite{yang2015automatic,liu2015automatic}.

In order to effectively incorporate context when predicting a label, the features for a given token are computed based on that token and on the four surrounding tokens.

\subsection{ANN model} \label{sec:ann-system}

 The main components of the ANN model are recurrent neural networks (RNNs). In particular, we use a type of RNN called Long Short Term Memory (LSTM)~\cite{hochreiter1997long}, as discussed in Section \ref{sec:lstm}. 

\begin{figure*}[!ht]
\vspace{-0.3cm}
  \centering
      \includegraphics[width=1.0\textwidth]{{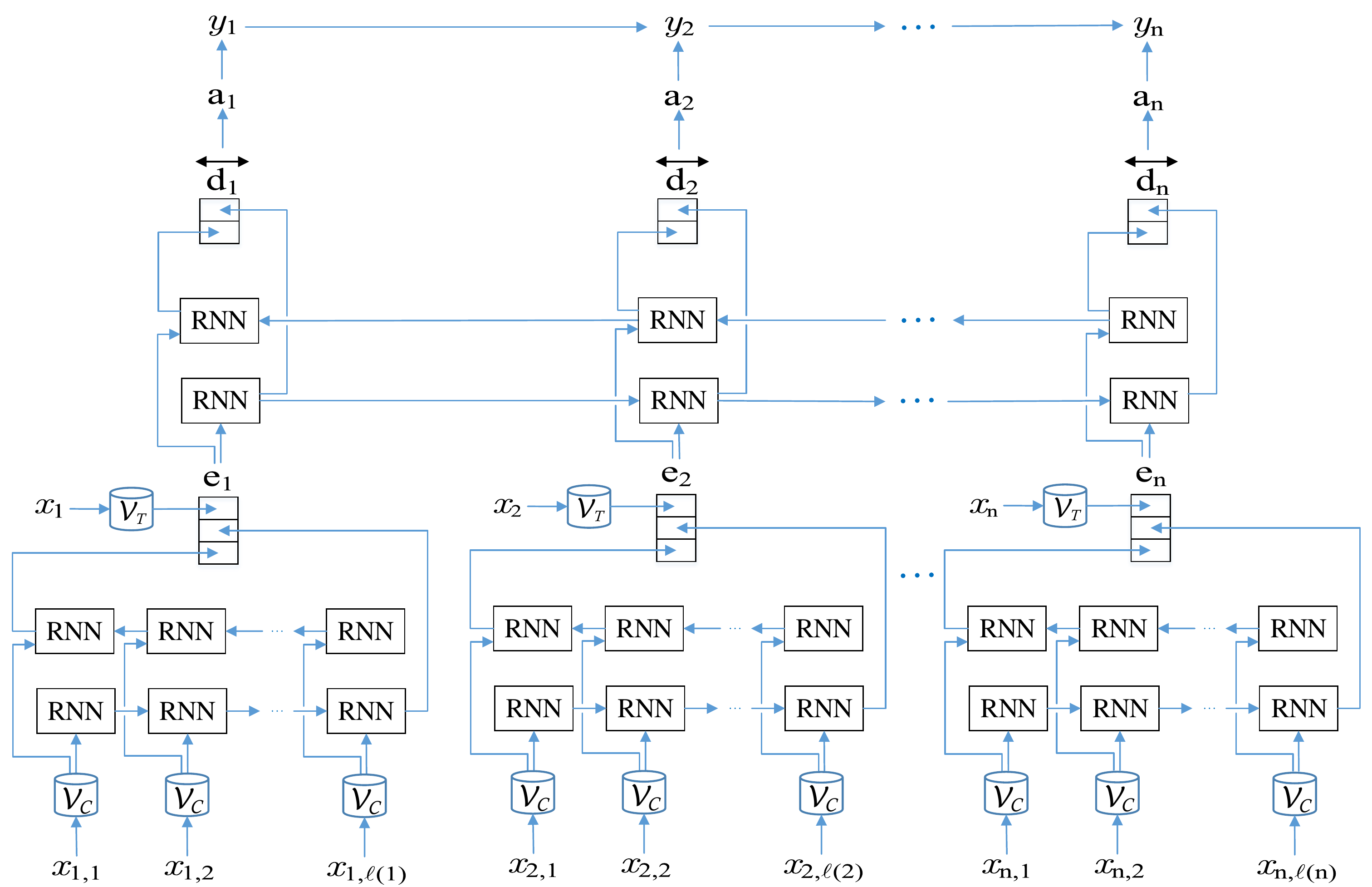}}
\vspace{-0.3cm}
  \caption{
  Architecture of the artificial neural network (ANN) model. RNN stands for recurrent neural network. The type of RNN used in this model is Long Short Term Memory (LSTM). $n$ is the number of tokens, and $x_i$ is the $i^{th}$ token. $\mathcal{V}_T$ is the mapping from tokens to token embeddings. $\ell(i)$ is the number of characters and $x_{i,j}$ is the $j^{th}$ character in the $i^{th}$ token. $\mathcal{V}_C$ is the mapping from characters to character embeddings. $\mathbf{e}_i$ is the character-enhanced token embeddings of the $i^{th}$ token. $\protect\overleftrightarrow{\mathbf{d}_i}$ is the output of the LSTM of label prediction layer, $\mathbf{a}_i$ is the probability vector over labels, $y_i$ is the predicted label of the $i$\textsuperscript{th} token. 
  }
  \vspace{-0.2cm}
  \label{fig:ann}
\end{figure*}

The system is composed of three layers: 
\vspace{-0.15cm} 
\begin{itemize}[noitemsep,leftmargin=*]
\item Character-enhanced token embedding layer (Section \ref{sec:embedding}),
\item Label prediction layer (Section \ref{sec:prediction}),
\item Label sequence optimization layer (Section \ref{sec:sequence-optimization}). 
\end{itemize} 
\vspace{-0.15cm} 
The character-enhanced token embedding layer maps each token into a vector representation. The sequence of vector representations corresponding to a sequence of tokens are input to the label prediction layer, which outputs the sequence of vectors containing the probability of each label for each corresponding token. Lastly, the sequence optimization layer outputs the most likely sequence of predicted labels based on the sequence of probability vectors from the previous layer. All layers are learned jointly. Figure~\ref{fig:ann} shows the ANN architecture.

In the following, we denote scalars in italic lowercase (e.g., $k$, $b_f$), vectors in bold lowercase (e.g., $\mathbf{s},\, \mathbf{x}_i$), and matrices in italic uppercase (e.g., $W_f$) symbols.
We use the colon notations $x_{i:j}$ and $ \mathbf{v}_{i:j}$ to denote the sequence of scalars $(x_i, \dotsc, x_j)$, and vectors $(\mathbf{v}_i, \mathbf{v}_{i+1}, \dotsc, \mathbf{v}_j)$, respectively.

\vspace{-0.05cm} 

\subsubsection{Bidirectional LSTM} \label{sec:lstm}
RNN is a neural network architecture designed to handle input sequences of variable sizes, but it fails to model long term dependencies. LSTM is a type of RNN that mitigates this issue by keeping a memory cell that serves as a summary of the preceding elements of an input sequence. More specifically, given a sequence of vectors $\mathbf{x}_{1}, \mathbf{x}_{2}, \dotsc, \mathbf{x}_{n},$ at each step $t = 1,\dotsc,n$, an LSTM takes as input $\mathbf{x}_{t}, \mathbf{h}_{t-1}, \mathbf{c}_{t-1}$ and produces the hidden state $\mathbf{h}_t$ and the memory cell $\mathbf{c}_t$ based on the following formulas:

\vspace{-0.55cm}
\begin{align*}
\mathbf{i}_t &= \sigma(W_i \; [ \mathbf{x}_{t};\; \mathbf{h}_{t-1};\; \mathbf{c}_{t-1} ] + \mathbf{b}_i) \\% \MoveEqLeft[2] \\
\mathbf{c}_t &=  ( 1 - \mathbf{i}_t ) \odot \mathbf{c}_{t-1} \\ 
	& \qquad { } + \mathbf{i}_t \odot \text{tanh}(W_c \; [ \mathbf{x}_{t};\; \mathbf{h}_{t-1}] + \mathbf{b}_c) ]\\
\mathbf{o}_t &= \sigma(W_o \; [ \mathbf{x}_{t};\; \mathbf{h}_{t-1};\; \mathbf{h}_{t-1} ] + \mathbf{b}_o) \\
\mathbf{h}_t &= \mathbf{o}_t \odot \text{tanh}(\mathbf{c}_t) 
\end{align*}
where $W_i, W_c, W_o$ are weight matrices and $ \mathbf{b}_i, \mathbf{b}_c,  \mathbf{b}_o$ are bias vectors used in the input gate, memory cell, and output gate calculations, respectively. 
\noindent The symbols $\sigma(\cdot)$ and tanh$(\cdot)$ refer to the element-wise sigmoid and hyperbolic tangent functions, and $\odot$ is the element-wise multiplication. $\mathbf{h}_0 = \mathbf{c}_0 = \mathbf{0}$.

A bidirectional LSTM consists of a forward LSTM and a backward LSTM, where the forward LSTM calculates the forward hidden states $(\overrightarrow{\mathbf{h}}_1, \overrightarrow{\mathbf{h}}_2, \dotsc, \overrightarrow{\mathbf{h}}_n),$ and the backward LSTM calculates the backward hidden states $(\overleftarrow{\mathbf{h}}_1, \overleftarrow{\mathbf{h}}_2, \dotsc, \overleftarrow{\mathbf{h}}_n)$ by feeding the input sequence in the backward order, from $\mathbf{x}_n$ to $\mathbf{x}_1.$ 

Depending on the application of the LSTM, one might need an output sequence corresponding to each element in the sequence, or a single output that summarizes the whole sequence. In the former case, the output sequence $\mathbf{h}_{1}, \mathbf{h}_{2}, \dotsc, \mathbf{h}_{n}$ of the LSTM is obtained by concatenating the hidden states of the forward and the backward LSTMs for each element i.e., $\overleftrightarrow{\mathbf{h}_t} = (\overrightarrow{\mathbf{h}}_t; \overleftarrow{\mathbf{h}}_t)$ for $t = 1,\dotsc,n.$ In the latter case, the output is obtained by concatenating the last hidden states of the forward and the backward LSTMs i.e., $\overleftrightarrow{\mathbf{h}} = (\overrightarrow{\mathbf{h}}_n; \overleftarrow{\mathbf{h}}_n).$

\subsubsection{Character-enhanced token embedding layer} \label{sec:embedding}

The character-enhanced token embedding layer takes a token as input and outputs its vector representation. The latter results from the concatenation of two different types of embeddings: the first one directly maps a token to a vector, while the second one comes from the output of a character-level token encoder. 

The direct mapping $\mathcal{V}_T(\cdot)$ from token to vector, often called a token (or word) embedding, can be pre-trained on large unlabeled datasets using programs such as word2vec~\cite{mikolov2013distributed,mikolov2013efficient,mikolov2013linguistic} or GloVe~\cite{pennington2014glove}, and can be learned jointly with the rest of the model. Token embeddings, often learned by sampling token co-occurrence distributions, have desirable properties such as locating semantically similar words closely in the vector space, hence leading to state-of-the-art performance for various tasks.  

While the token embeddings capture the semantics of tokens to some degree, they may still suffer from data sparsity. For example, they cannot account for out-of-vocabulary tokens, misspellings, and different noun forms or verb endings. One solution to remediate some of these issues would be to lemmatize tokens before training, but this approach may fail to retain some useful information such as the distinction between some verb and noun forms. 

We address this issue by using character-based token embeddings, which incorporate each individual character of a token to generate its vector representation. This approach enables the model to learn sub-token patterns such as morphemes (e.g., suffix or prefix) and roots, thereby capturing out-of-vocabulary tokens, different surface forms, and other information not contained in the token embeddings.

Let $x_{i,1},\dotsc,x_{i,\ell(i)}$ be the sequence of characters that comprise the $i^{th}$ token $x_i$, where $\ell(i)$ is the number of characters in $x_i.$ The character-level token encoder generates the character-based token embedding of $x_i$ by first mapping each character $x_{i,j}$ to a vector $\mathcal{V}_C(x_{i,j}),$ called a character embedding, via the mapping $\mathcal{V}_C(\cdot).$ Then the sequence $\mathcal{V}_C(x_{i,j})$ is passed to a bidirectional LSTM, which outputs the character-based token embedding $\overleftrightarrow{\mathbf{b}_i}$

As a result, the final output $\mathbf{e}_{i}$ of the character-enhanced token embedding layer for $i^{th}$ token $x_{i}$ is the concatenation of the token embedding $\mathcal{V}_T(x_{i})$ and the character-based token embedding $\overleftrightarrow{\mathbf{b}_i}.$ 
In summary, when the character-enhanced token embedding layer receives a sequence of tokens $x_{1:n}$ as input, it will output the sequence of token embeddings $\mathbf{e}_{1:n}$. 

\subsubsection{Label prediction layer} \label{sec:prediction}

The label prediction layer takes as input the sequence of vectors $\mathbf{e}_{1:n}$, i.e., the outputs of the character-enhanced token embedding layer, and outputs $\mathbf{a}_{1:n}$, where the $t^{th}$ element of $\mathbf{a}_n$ is the probability that the $n^{th}$ token has the label $t$. The labels are either one of the PHI types or non-PHI. For example, if one aims to predict all 18 HIPAA-defined PHI types, there would be 19 different labels.

The label prediction layer contains a bidirectional LSTM that takes the input sequence $\mathbf{e}_{1:n}$ and generates the corresponding output sequence $\overleftrightarrow{\mathbf{d}_{1:n}}.$ Each output $\overleftrightarrow{\mathbf{d}_i}$ of the LSTM is given to a feed-forward neural network with one hidden layer, which outputs the corresponding probability vector $\mathbf{a}_{i}$.

\subsubsection{Label sequence optimization layer} \label{sec:sequence-optimization}

The label sequence optimization layer takes the sequence of probability vectors $\mathbf{a}_{1:n}$ from the label prediction layer as input, and outputs a sequence of labels $y_{1:n}$, where $y_{i}$ is the label assigned to the token $t_{i}$.

The simplest strategy to select the label $y_{i}$ would be to choose the label that has the highest probability in $\mathbf{a}_{i}$, i.e. $y_{i}=\argmax_{k}{\mathbf{a}_{i}[k]}$. However, this greedy approach fails to take into account the dependencies between subsequent labels. For example, it may be more likely to have a token with the PHI type STATE followed by a token with the PHI type ZIP than any other PHI type. Even though the label prediction layer has the capacity to capture such dependencies to a certain degree, it may be preferable to allow the model to directly learn these dependencies in the last layer of the model.

One way to model such dependencies is to incorporate a matrix $T$ that contains the transition probabilities between two subsequent labels. $T[i,j]$ is the probability that a token with label $i$ is followed by a token with the label $j$. The score of a label sequence $y_{1:n}$ is defined as the sum of the probabilities of individual labels and the transition probabilities:
$$ s(y_{1:n}) = { \sum_{i=1}^{n} \mathbf{a}_{i}[y_{i}]+  \sum_{i=2}^{n} T [y_{i-1},y_{i}} ]. $$
These scores can be turned into probabilities of the label sequences by taking a softmax function over all possible label sequences.  During the training phase, the objective is to maximize the log probability of the gold label sequence. In the testing phase, given an input sequence of tokens, the corresponding sequence of predicted labels is chosen as the one that maximizes the score.

%% file: results.tex
\section{Experiments and results} \label{sec:results}

\subsection{Datasets}

We evaluate our two models on two datasets: i2b2 2014 and MIMIC de-identification datasets. The i2b2 2014 dataset was released as part of the 2014 i2b2/UTHealth shared task Track~1~\cite{stubbs2015automated}. It is the largest publicly available dataset for de-identification. Ten teams participated in this shared task, and 22 systems were submitted. As a result, we used the i2b2 2014 dataset to compare our models against state-of-the-art systems.

The MIMIC de-identification dataset was created for this work as follows. The MIMIC-III dataset~\cite{mimic3,goldberger2000physiobank,saeed2011multiparameter} contains data for 61,532 ICU stays over 58,976 hospital admissions for 46,520 patients, including 2 million patient notes. In order to make the notes publicly available, a rule-based de-identification system~\cite{douglass2005computer,douglass2005identification,douglas2004computer} was written for the specific purpose of de-identifying patient notes in MIMIC, leveraging dataset-specific information such as the list of patient names or addresses. The system favors recall over precision: there are virtually no false negatives, while there are numerous false positives. To create the gold standard MIMIC de-identification dataset, we selected 1,635 discharge summaries, each belonging to a different patient, containing a total of 60.7k PHI instances. We then annotated the PHI instances detected by the rule-based system as true positives or false positives. We found that 15\% of the PHI instances detected by the rule-based system were false positives. 

Table~\ref{tab:phi_types} introduces the PHI types and Table~\ref{tab:datasets} presents the datasets' sizes. For the test set, we used the official test set for the i2b2 dataset, which is 40\% of the dataset; we randomly selected 20\% of the MIMIC dataset as the test set for this dataset.

\input{tables/datasets}

\subsection{Evaluation metrics}

To assess the performance of the two models, we computed the precision,
recall, and F1-score. Let TP be the number of true positives, FP the number of false positives, and FN the number of false negatives. Precision, recall, and F1-score are defined as follows: $\text{precision}=\frac{TP}{TP +FP}$,  $\text{recall}=\frac{TP}{TP +FN}$, and $\text{F1-score}=\frac{2*\text{precision}*\text{recall}}{\text{precision} +\text{recall}}$. Intuitively, precision is the proportion of the predicted PHI labels that are gold labels, recall is the proportion of the gold PHI labels that are correctly predicted, and F1-score is the harmonic mean of precision and recall.

\subsection{Training and hyperparameters}
The model is trained using stochastic gradient descent, updating all parameters, i.e., token embeddings, character embeddings, parameters of bidirectional LSTMs, and transition probabilities, at each gradient step. For regularization, dropout is applied to the character-enhanced token embeddings before the label prediction layer. Below are the choices of hyperparameters and token embeddings, optimized using a subset of the training set:
\begin{itemize}[noitemsep]
\item character embedding dimension: 25
\item character-based token embedding LSTM dimension: 25
\item token embedding dimension: 100
\item label prediction LSTM dimension: 100
\item dropout probability: 0.5
\end{itemize}

We tried pre-training token embeddings on the i2b2 2014 dataset and the MIMIC dataset\footnote{For MIMIC, we used the entire dataset containing 2 million notes and 800 million tokens.} using word2vec and GloVe. Both word2vec and GloVe were trained using a window size of 10, a minimum vocabulary count of 5, and 15 iterations. Additional parameters of word2vec were the negative sampling and the model type, which were set to 10 and skip-gram, respectively. We also experimented with the publicly available\footnote{\url{http://nlp.stanford.edu/projects/glove/}} token embeddings such as GloVe trained on Wikipedia and Gigaword 5~\cite{parker2011english}. The results were quite robust to the choice of the pre-trained token embeddings. The GloVe embeddings trained on Wikipedia articles yielded slightly better results, and we chose them for the rest of this work.

\input{tables/main_result}

\begin{figure*}[!ht]
  \centering

            \includegraphics[width=0.45\textwidth]{{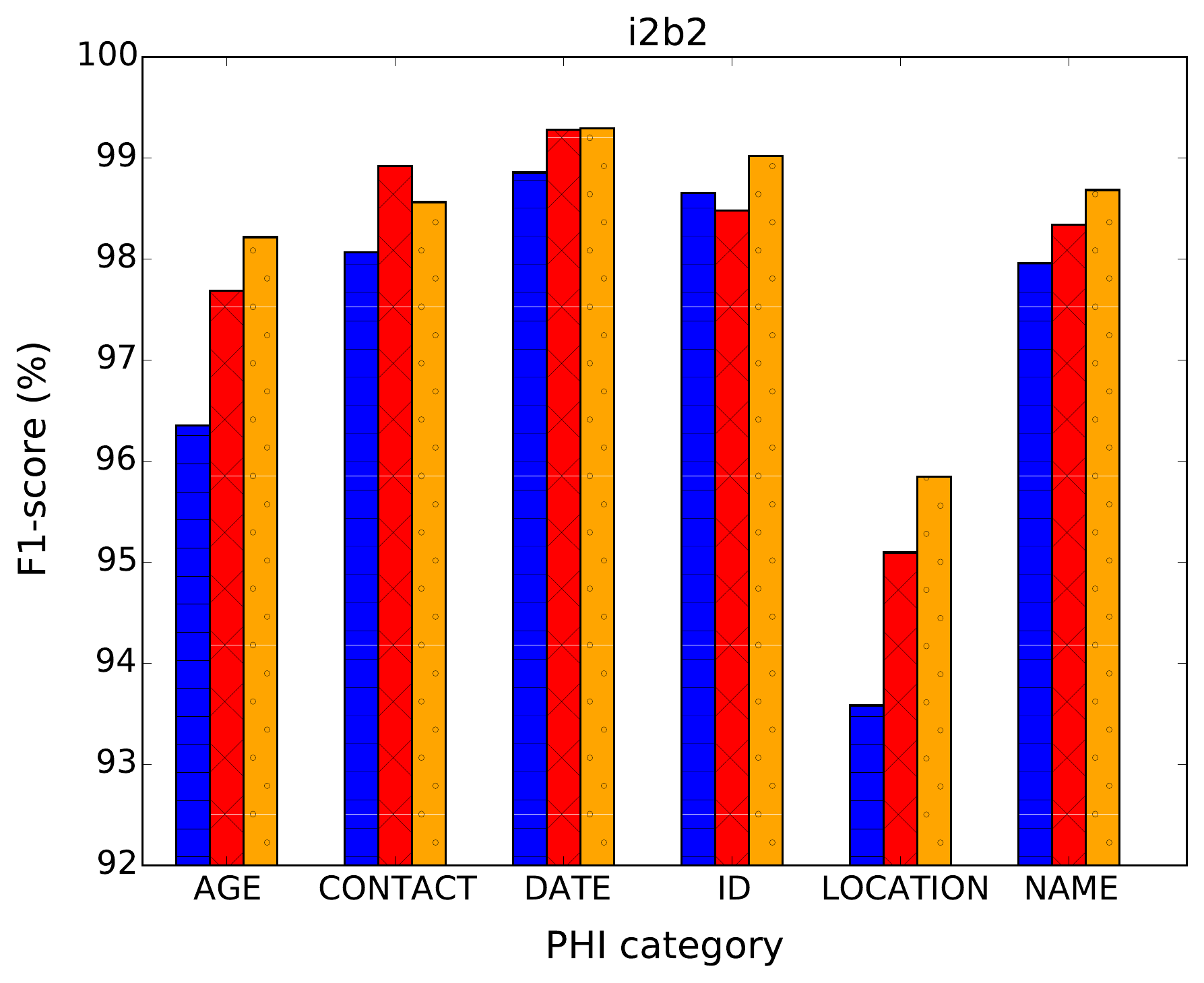}}
            \hspace{1cm}
      \includegraphics[width=0.45\textwidth]{{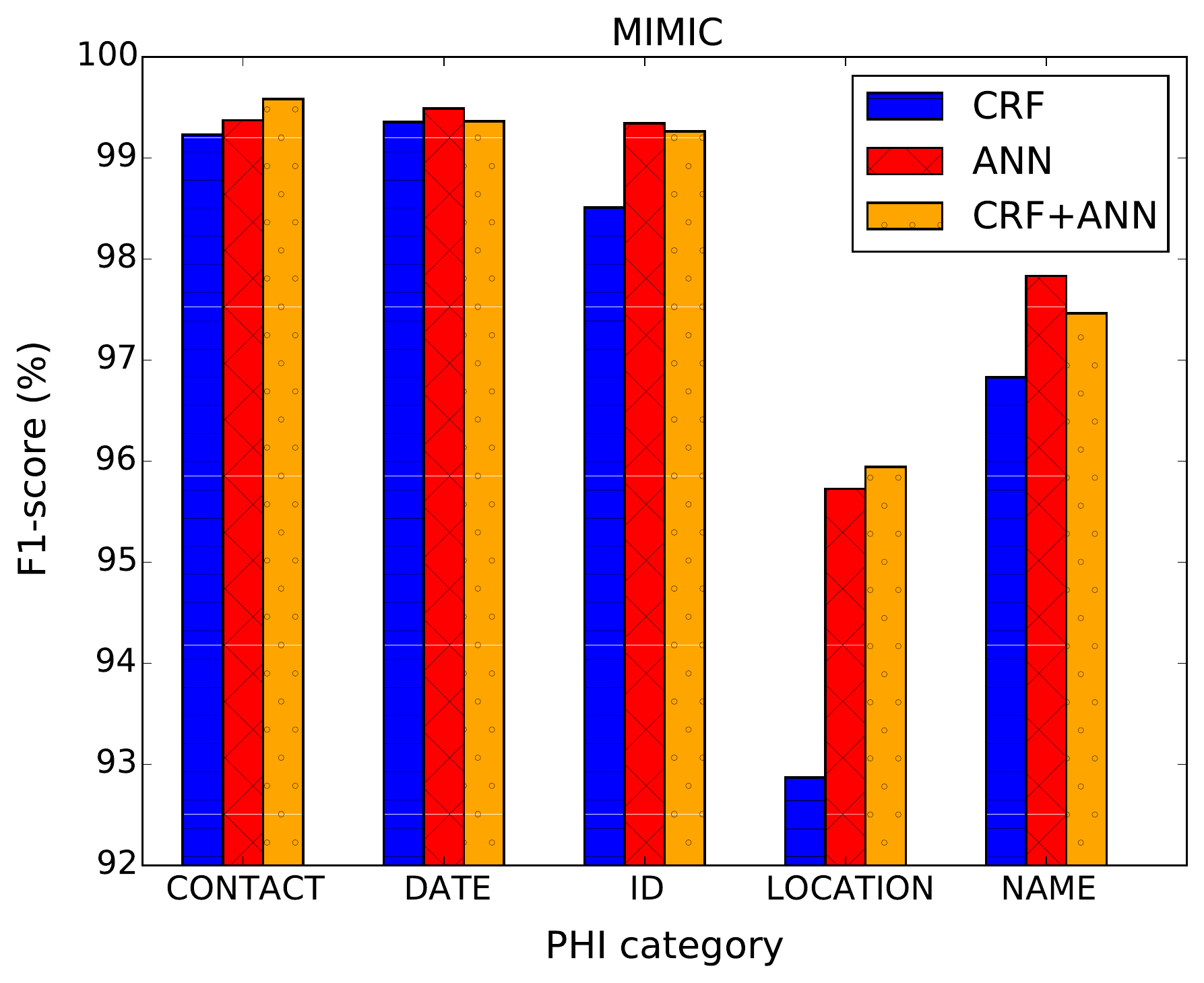}}
\vspace{0.0cm}

  \caption{Binary token-based F1-scores for each PHI category. The evaluation is based on PHI types that are defined by HIPAA as well as additional PHI types specific to each dataset. Each PHI category and the corresponding PHI types are defined in Table 1. The PROFESSION category exists only in the i2b2 dataset, and was removed from the graph to avoid distorting the y-axis: the F1-scores are 72.014, 82.035, and 81.664 with the CRF, ANN, and CRF+ANN, respectively. For the same reason, the AGE category in MIMIC was removed: the F1-scores are 80.851, 81.481, and 92.308 with the CRF, ANN, and CRF+ANN, respectively.} 
  \vspace{-0.3cm}
  \label{tab:per_category}
\end{figure*}

\subsection{Results}

All results were computed using the official evaluation script from the i2b2 2014 de-identification challenge.
Table~\ref{tab:main_result} presents the main results, based on binary token-based precision, recall, and F1-score for HIPAA-defined PHI only. These PHI types are the most important since only those are required to be removed by law. On the i2b2 dataset, our ANN model has a higher F1-score and recall than our CRF model as well as the best system from the i2b2 2014 de-identification challenge, which was the Nottingham system~\cite{yang2015automatic}. The only freely available, off-the-shelf program for de-identification, called the MITRE Identification Scrubber Toolkit (MIST)~\cite{aberdeen2010mitre}, performed poorly. Combining the outputs of our ANN and CRF models, by considering a token to be PHI if it is identified as such by either model, further increases the performance in terms of F1-score and recall.

It should be noted that the Nottingham system was specifically fine-tuned for the i2b2 dataset as well as the i2b2 evaluation script. For example, the Nottingham system post-processes the detected PHI terms in order to match the offset of the gold PHI tokens, such as modifying ``MR:6746781" to ``6746782" and ``MWFS" to ``M", ``W", ``F", ``S".

On the MIMIC dataset, our ANN model also has a higher F1-score and recall than our CRF model. Interestingly, combining the outputs of our ANN and CRF models did not increase the F1-score, because precision was negatively impacted. However, the recall did benefit from combining the two models. MIST was much more competitive on this dataset.

We calculated the statistical significance of the differences in precision, recall, and F1-score between the CRF and ANN models using approximate randomization with 9999 shuffles. The significance levels of the differences in precision, recall, and F1-score are 0.37, 0.02, 0.22 for the i2b2 dataset, and 0.08, 0.00, 0.00 for the MIMIC dataset, respectively.

\input{tables/analysis}

\subsection{Error analysis}

Figure~\ref{tab:per_category} shows the binary token-based F1-scores for each PHI category. The ANN model outperforms the CRF model on all categories for both datasets, with the exception of the ID (which mostly contains medical record numbers) category in the i2b2 dataset. This is due to the fact that the CRF model uses sophisticated regular expression features that are tailored to detect ID patterns such as ``38:Z8912708G''. 

Another interesting difference between the ANN and the CRF results is the PROFESSION category: the ANN significantly outperforms the CRF. The reason behind this result is that the embeddings of the tokens that represent a profession tend to be close in the token embedding space, which allows the ANN model to generalize well. We tried assembling various gazetteers for the PROFESSION category, but all of them were performing significantly worse than the ANN model.

Table~\ref{tab:analysis} presents some examples of gold PHI instances correctly predicted by the ANN model that the CRF model failed to predict, and conversely. This illustrates that the ANN model efficiently copes with the diversity of the contexts in which tokens appear, whereas the CRF model can only address the contexts that are manually encoded as features. In other words, the ANN model's intrinsic flexibility allows it to better capture the variance in human languages than the CRF model. For example, it would be challenging and time-consuming to engineer features for all possible contexts such as ``had a stroke at 80'', ``quit smoking in 08'', ``on the 29th of this month'', and ``his friend Epstein''.  The ANN model is also very robust to variations in surface forms, such as misspellings (e.g., ``in teh late 60s'', ``Khazakhstani'', ``01/19/:0''), tokenizations (e.g., ``Results02/20/2087'', ``MC \# 0937884Date''), and different phrases referring to the same semantic meaning (e.g., ``San Rafael Mount Hospital'', ``Rafael Mount'', ``Rafael Hospital''). Furthermore, the ANN model is able to detect many PHI instances despite not having explicit gazetteers, as examples in the LOCATION and PROFESSION categories illustrate. We conjecture that the character-enhanced token embeddings contain rich enough information to effectively function as gazetteers, as tokens with similar semantics are closely located in the vector representation~\cite{mikolov2013distributed,collobert2011natural,kim2015character}.

On the other hand, CRF is good at rarely occurring patterns that are written in highly specialized regular expression patterns (e.g., ``38:Z8912708G'', ``53RHM'') or tokens that are included in the gazetteers (e.g., ``Christmas'', ``WPH'', ``rosenberg'', ``Motor Vehicle Body Repairer''). For example, the PHI token ``Christmas'' only occurs in the test set, and unless the context gives a strong indication, the ANN model cannot detect it, whereas the CRF model could, as long as it is included in the gazetteers.

\subsection{Effect of training set size}

\begin{figure}[!ht]
  \centering
      \includegraphics[width=0.45\textwidth]{{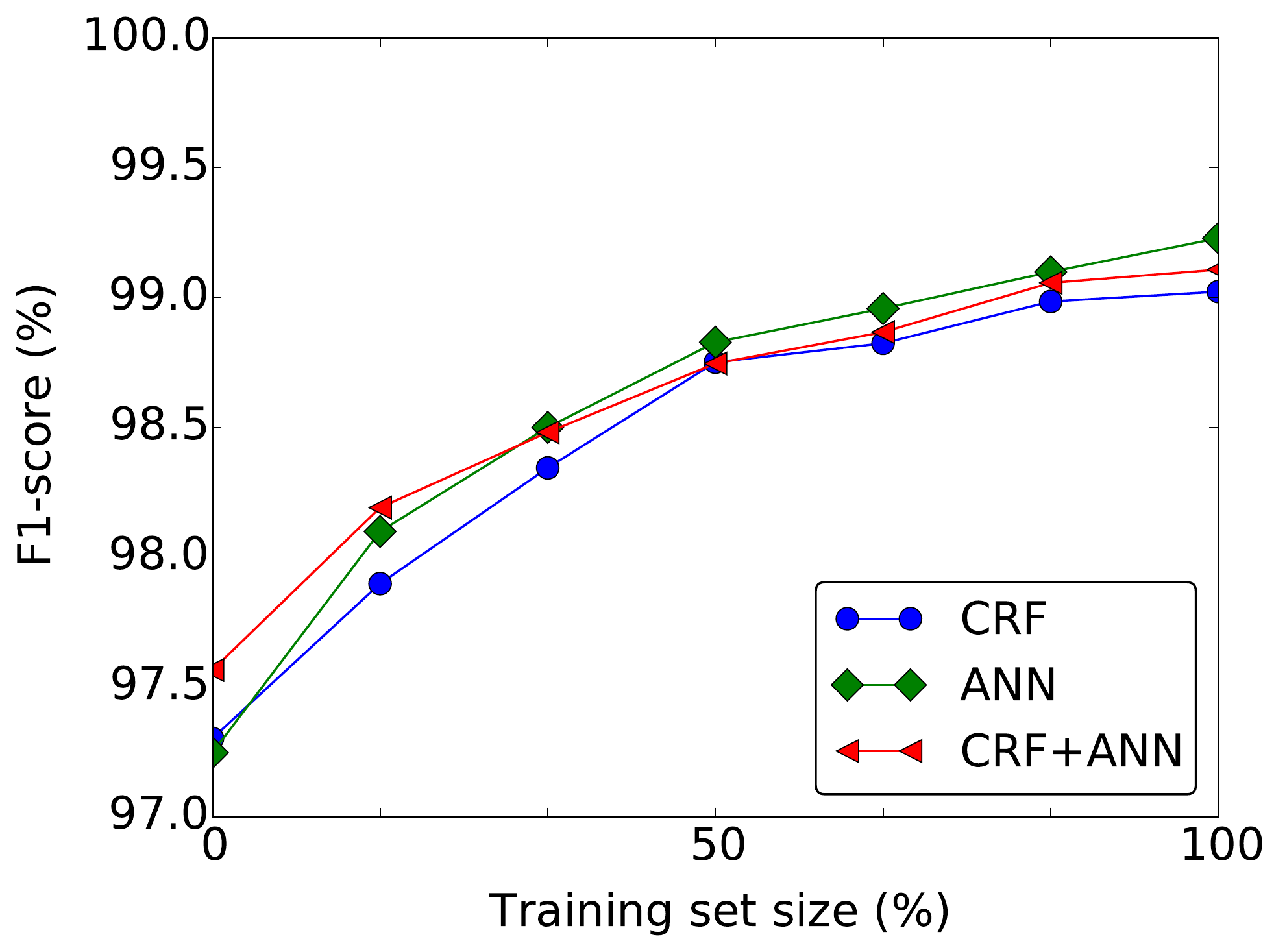}}
  \caption{Impact of the training set size on the binary HIPAA token-based F1-scores on the MIMIC dataset. 100\% training set size refers to using all of the dataset minus the test set.} 
  \label{tab:training_set_size}
\end{figure}

\begin{figure}[!ht]
  \centering

      \includegraphics[width=0.45\textwidth]{{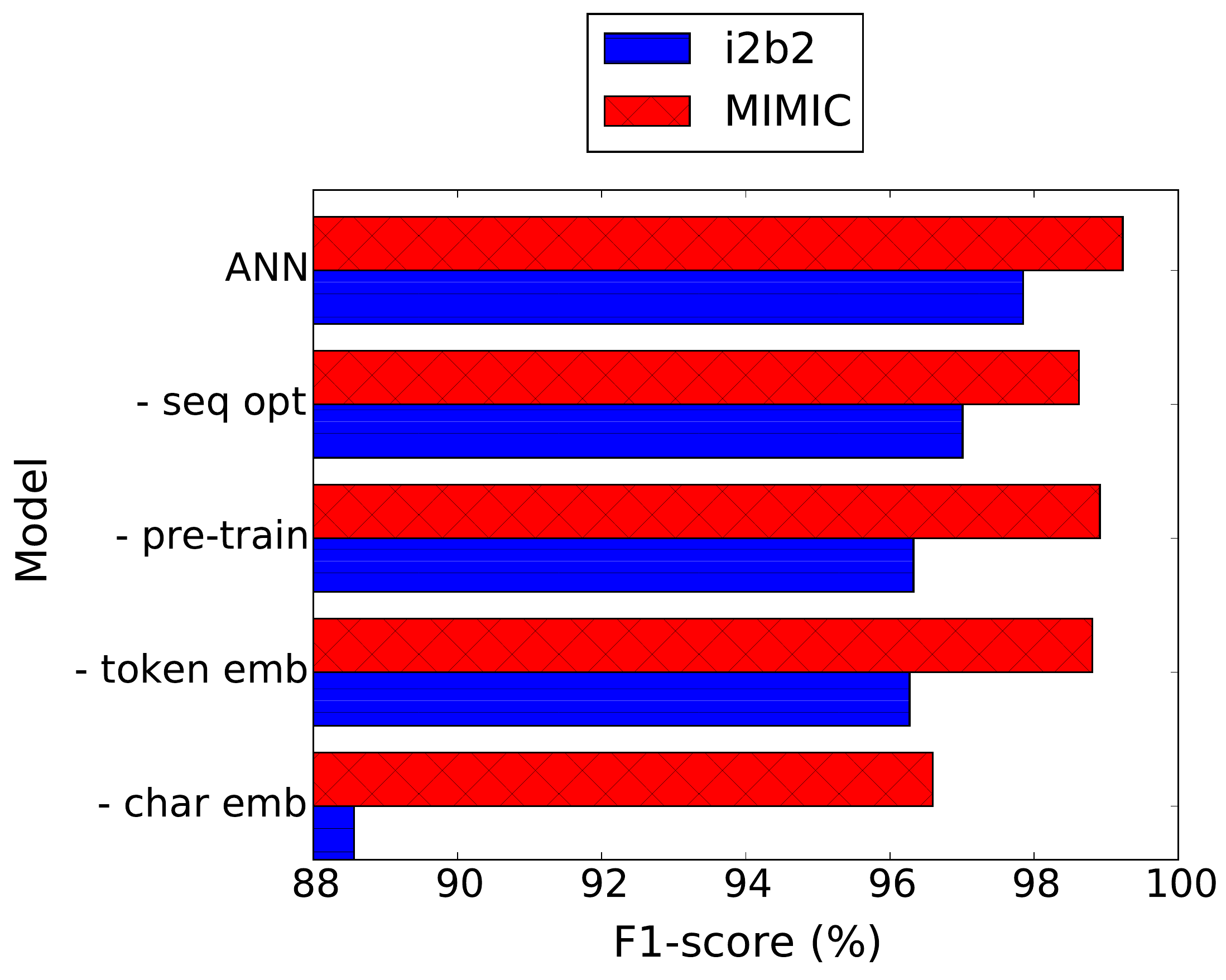}}
  \caption{Ablation test performance based on binary HIPAA token-based evaluation. ANN is the model based on Artificial Neural Network. “- seq opt” is the ANN model without the label sequence optimization layer. “- pre-train” is the ANN model where token embeddings are initialized with random values instead of pre-trained embeddings. “- token emb” is the ANN model using only character-based token embeddings, without token embeddings. “- character emb” is the ANN model using only token embeddings, without character-based token embeddings.}
  \label{tab:ablation}
\end{figure}

Figure~\ref{tab:training_set_size} shows the impact of the training set size on the performance of the models on the MIMIC dataset. When the training set size is very limited, the CRF performs slightly better than the ANN model, since the CRF model can leverage handcrafted features without much training data. As the training set size increases, the ANN model starts to significantly outperform the CRF model, since the parameters including the embeddings are automatically fine-tuned with more data, and therefore the features learned by the ANN model become increasingly more refined than the manually handcrafted features. As a result, combining the outputs of the CRF and ANN models increases the F1-score over the ANN model only for small training set size and yields a less competitive F1-score than the ANN model for bigger training set size.

\subsection{Ablation analysis}
\vspace{0.2cm}
In order to quantify the importance of various elements of the ANN model, we tried 4 variations of the model, eliminating different elements one at a time. Figure~\ref{tab:ablation} presents the results of the ablation tests. Removing either the label sequence optimization layer, pre-trained token embeddings, or token embeddings slightly decreased the performance. Surprisingly, the ANN performed pretty well with only character embeddings and without the token embeddings, and eliminating the character embeddings was more detrimental than eliminating the token embeddings. This suggests that the character-based token embeddings may be capturing not only the sub-token level features, but also the semantics of the tokens themselves.

%% file: tables/datasets.tex
\begin{table} [H]
\footnotesize
\centering
\setlength\tabcolsep{6.0pt}
\setlength{\extrarowheight}{3pt}
\setlength{\arraycolsep}{5pt}
\begin{tabular}{|l|c|c|}
\hline
\textbf{} & \textbf{i2b2} 	& \textbf{MIMIC} \\
\hline
\text{Vocabulary size}	& 46,803	&	69,525\\ 
\text{Number of notes}	&	1,304& 	1,635\\
\text{Number of tokens}	&	984,723& 2,945,228	\\  
\text{Number of PHIs}	&	28,867	& 60,725\\  
\text{Number of PHI tokens}	&41,355	& 78,633	\\  
\hline
\end{tabular}
\caption{Overview of the i2b2 and MIMIC datasets.
} \label{tab:datasets}
\end{table}

%% file: tables/main_result.tex
 \begin{table*}[!ht]
\footnotesize
\vspace{-0.1cm}
\centering
\setlength\tabcolsep{6.0pt}
\setlength{\extrarowheight}{3pt}
\setlength{\arraycolsep}{5pt}
\begin{tabular}{|c|ccc|ccc|}
\hline 
\multirow{2}{*}{Model} & \multicolumn{3}{c|}{i2b2} & \multicolumn{3}{c|}{MIMIC}\tabularnewline
\cline{2-7} 
 & Precision & Recall & F1-score & Precision & Recall & F1-score\tabularnewline
\hline 
Nottingham & \textbf{99.000 }& 96.680 & 97.680 & - & - & -\tabularnewline
MIST & 95.288  & 75.691 & 84.367 & 97.739 & 97.164 & 97.450\tabularnewline
CRF & 98.560 & 96.528 & 97.533 & 99.060 & 98.987 & 99.023\tabularnewline
ANN & 98.320 & 97.380 & 97.848 & \textbf{99.208} & 99.251 & \textbf{99.229}\tabularnewline
CRF + ANN & 97.920 & \textbf{97.835} & \textbf{97.877} & 98.820 & \textbf{99.398}& 99.108\tabularnewline
\hline 
\end{tabular}
\caption{Performance (\%) on the 
PHI 
as defined in the 
HIPAA. We evaluated the systems based on the detection of PHI token versus non-PHI token (i.e., binary HIPAA token-based evaluation). The best performance for each metric on each dataset is highlighted in bold. Nottingham is the best performing system from the 2014 i2b2/UTHealth shared task Track 1. MIST, the MITRE Identification Scrubber Toolkit, is a freely available de-identification program. CRF is the model based on Conditional Random Field, ANN is the model based on Artificial Neural Network, and CRF+ANN is the result obtained by combining the outputs of the CRF model and the ANN model. The Nottingham system could not be run on the MIMIC dataset, as it is not publicly available. } \label{tab:main_result}
\end{table*}

%% file: tables/analysis.tex
\begin{table*}[!ht]
\scriptsize
\centering
\setlength\tabcolsep{4.0pt}
\setlength{\extrarowheight}{3pt}
\setlength{\arraycolsep}{5pt}
\begin{center}
\begin{adjustbox}{center}
\begin{tabular}{|c|c|c|c|c|c|}
\hline
\textbf{PHI category} & ANN 	& CRF	 \\
\hline
AGE 	& Father had a stroke at \textbf{80} and died of ?another stroke at age 83.	&	HPI:  \textbf{53}RHM who going to bed Wednesday was in usoh, but 	 	\\
	&  PERSONAL DATA AND OVERALL HEALTH:  Now \textbf{63}, despite his &	 	Tobacco: Quit at \textbf{38} y/o; ETOH: 1-2 beers/week; Caffeine: \\
	&   FH: Father: Died @ \textbf{52} from EtOH abuse (unclear exact etiology)      &        \\
		&  Tobacco: smoked from age 7 to \textbf{15}, has not smoked since 15.       &        \\
			&   History of Present Illness  \textbf{86}F reports worsening b/l leg pain.       &        \\\hline
CONTACT 				&     by phone, Dr. Ivan Guy.   Call w/ questions \textbf{86383}.    Keith Gilbert,     &        \\
&   H/O paroxysmal afib  VNA \textbf{171-311-7974}  ======= Medications      &        \\\hline
DATE &    During his \textbf{May} hospitalization he had dysphagia      &    She is looking forward to a good \textbf{Christmas}. She is here today    \\
&     Social history:  divorced, quit smoking in \textbf{08}, sober x 10 yrs,    &        \\
&      She is to see him on the \textbf{29th} of this month at 1:00 p.m.     &        \\
&      He did have a renal biopsy in teh late \textbf{60s} adn thus will look for results,    &        \\
&   Results\textbf{02/20/2087} NA 135,  K 3.2 (L),  CL 96 (L),  CO2 30.6,  BUN 1      &        \\
&      Jose Church, M.D.  /ray  DD: 01/18/20  DT: \textbf{01/19/:0}  DV: 01/18/20   &        \\\hline
ID &     placed 3/23 for bradycardia. P/G model \# \textbf{5435}, serial \# 4712198,     &       DD:05/05/2095 DT:05/05/2095 \textbf{WK:65255     :4653} \\
&     Consult NotePt: Ulysses Ogrady  MC \# \textbf{0937884}Date: 10/07/69    &  NO GROWTH TO DATE   Specimen: \textbf{38:Z8912708G}   Collected       \\\hline
LOCATION &   Works in programming at \textbf{Audiovox}.  Formerly at BrightPoint.      &  2nd set biomarkers (\textbf{WPH}): Creatine Kinase Isoenzymes         \\
&        He has remote travel hx to the \textbf{Rockefeller Centre}, more recent global  &   Hospitalized 2115 \textbf{TCH} for ROMI  2120 TCH new onset     \\
&      History of Present Illness:  Pt is a 59 yo \textbf{Khazakhstani} male, with    &        \\
&      who was admitted to \textbf{San Rafael Mount Hospital }following a syncopal    &        \\
&   nauseas and was brought to \textbf{Rafael Mount }ED.  Five weeks ago prior       &        \\
&   Anemia: On admission to \textbf{Rafael Hospital}, Hb/Hct: 11.6/35.5.      &        \\\hline
NAME &     ATCH:  655-75-45   Dear Harry and \textbf{Yair}:   My thanks for your kind    &  Lab Tests  \textbf{Amador}:  the lab results show good levels of      \\
&     Patient lives in Flint with his friend \textbf{Epstein}. He has 3 children.   &  10MG PO qd : 05/10/2066 - 04/15/2068 ACT : \textbf{rosenberg}      \\
&    Health care proxy-Yes, son (\textbf{West})   Allergies  DUTASTERIDE  - cough,      &   128 Williams Ct       M \textbf{OSCAR, JOHNNY}     Hyderabad, WI  62297        \\\hline
PROFESSION &   Social history:   Married, \textbf{glazier}, 3 grown adult children       &  Social history:   He is retried \textbf{Motor Vehicle Body Repairer}.       \\
&   Has VNA.  Former civil engineer, \textbf{supervisor}, consultant.       &        \\
&    He was formerly self-employed as a \textbf{CPA} and would often travel       &        \\
&     Communications senior manager, \textbf{marketing}, worked for Brinker      &        \\
&      and Concrete Finisher (25yrs). He is a \textbf{veteran}.   &        \\
&    Former tobacco user, works part time in \textbf{securities}.      &        \\
\hline
\end{tabular}
\end{adjustbox}
\end{center}
\vspace{0.1cm}
\caption{Examples of correctly detected PHI instances (in bold) by the ANN and CRF models for the i2b2 dataset. The examples in the ANN column are only predicted by the ANN model and not predicted by the CRF model, and conversely. Typographical errors are from the original text.} \label{tab:analysis}
\vspace{-0.3cm}
\end{table*}

%% file: conclusion.tex
\vspace{0.5cm}
\section{Conclusions}
\vspace{0.2cm}
We proposed the first system based on ANN for patient note de-identification. It outperforms state-of-the-art systems based on CRF on two datasets, while requiring no handcrafted features. Utilizing both the token and character embeddings, the system can automatically learn effective features from data by fine-tuning the parameters. It jointly learns the parameters for the embeddings, the bidirectional LSTMs as well as the label sequence optimization, and can make use of token embeddings pre-trained on large unlabeled datasets. Quantitative and qualitative analysis of the ANN and CRF models indicates that the ANN model better incorporates context and is more flexible to variations inherent in human languages than the CRF model. 

From the viewpoint of deploying an off-the-shelf de-identification system, our results in Table~\ref{tab:main_result} demonstrate recall on the MIMIC discharge summaries over 99\%, which is quite encouraging.  Figure~\ref{tab:per_category}, however, shows that the F1-score on the NAME category, probably the most sensitive PHI type, falls just below 98\% for the ANN model. We anticipate that adding gazetteer features based on the local institution's patient and staff census should improve this result, which will be explored in future work.